# Positive Ratings, Hidden Concerns: Employee Voice Disclosure in AI-Mediated Organizational Listening

| Thilo Tamme | Michael Saatkamp | Alma Bonte | Daniel Weiss | Anton Hantel | Andrej Levin |
|---|---|---|---|---|---|
| TUM | TUM | TUM | LMU | MIT | TUM |
| Thilo.tamme @tum.de | Michael.saatkamp @tum.de | Alma.bonte @tum.de | D.weiss @campus.lmu.de | Hantel @mit.edu | Andrej.levin @tum.de |

## Abstract

*Organizations started listening to employees through conversational AI agents alongside structured surveys. Little is known about what these channels change in what employees say when disclosure carries hierarchical risk. We report a field study inside a global management consulting firm whose process pairs a pre-survey with an adaptive AI voice interview on the same themes within one session. Across 44 first-session interviews (132 matched theme observations), 20–41% of sessions showed a favorable rating co-occurring with a substantive concern voiced later, depending on the favorability threshold. The Gioia analysis drew on 158 protective quotes from 65 eligible sessions. Disclosure rarely arrived unguarded: employees softened concerns, deflected accountability, and bounded how far they went, and this protective work tracked the perceived legitimacy of the listening structure. We develop a grounded model of bounded disclosure and derive four propositions for voice, channel and listening research. Silence, we argue, can persist inside expression.*

**Keywords:** Employee voice and silence, Conversational AI, Organizational listening, Self-disclosure

## 1. Introduction

Employees know things their organizations need to hear: which teams are burning out, which projects are quietly failing, which leaders have stopped listening. Much of this knowledge never travels upward. Decades of voice research show that employees withhold concerns that implicate those above them, and because silence is the absence of expression, it leaves no trace in the instruments organizations use to listen (Morrison & Milliken, 2000; Milliken et al., 2003; Morrison, 2023). The dominant instrument for organizational listening, the structured survey, may even compound the problem: a favorable rating is read as the absence of a concern, when it may merely be a safe place to keep one.

Conversational AI agents that conduct adaptive interviews at scale are now being deployed alongside surveys as a new channel of employee listening (Dutta et al., 2023; Wuttke et al., 2025). Existing research offers grounds for both optimism and doubt about whether they reach what surveys miss. Laboratory and consumer studies find that people disclose sensitive information willingly to conversational agents, in part because a non-human interlocutor is perceived as less evaluative (Ho et al., 2018; Lucas et al., 2014; Papneja & Yadav, 2025), and conversational interviewing yields more elaborated answers than fixed response formats (Schober & Conrad, 1997). Almost none of this evidence, however, comes from settings in which disclosure carries real hierarchical risk for the speaker. Missing is field evidence that observes the same employee, theme, and week across both channels, so the two renderings of one concern can be compared directly.

This study provides such evidence and asks: **When and how do employees disclose hierarchy-sensitive concerns in an AI conversation while giving favorable survey ratings on the same theme?** We address the question in a field study inside a global management consulting firm whose weekly listening process pairs a structured pre-survey with an adaptive voice interview conducted by a conversational AI agent, both channels covering the same themes within a single session. Analyzing 65 interview sessions and 158 protective quotes with the Gioia methodology for inductive theory building (Gioia et al., 2013), we develop a grounded model of how candor is managed across the two channels and derive four propositions.

The divergence between the channels proved substantial. Disclosure-gap prevalence ranged from 20% under a stricter favorability threshold to 41% under our primary operationalization: in these sessions, at least one favorable rating co-occurred with a substantive concern voiced on the same theme minutes later, such as an employee rating workload favorably and then describing working past midnight to the agent in the same session. Disclosure in the conversational channel rarely arrived unguarded: employees softened their concerns, deflected accountability away from superiors, and bounded how far a concern was taken, and how far this protective work relaxed tracked the perceived legitimacy of the listening structure.

The study makes three contributions, developed in §6.1. It identifies a second locus of silence, operating *inside* expression, where a concern is voiced yet rendered below the threshold a co-occurring rating registers (Morrison, 2014). It brings within-person field evidence that the disclosure-enhancing pattern of conversational AI survives real hierarchical stakes, but only as *managed* disclosure. And it separates two achievements that channel designs conflate: eliciting voice, which adaptive probing accomplishes, and earning trust in the listening structure, which only felt legitimacy sustains (Yip & Fisher, 2022).

The study speaks directly to *AI and the Future of Work*. It examines a socio-technical artifact, an AI listening agent, that reshapes a core collaborative practice: how organizations sense and respond to their workforce. Its central design lesson is that two system properties usually bundled together, the capacity to elicit candid input and the capacity to make that input consequential, are separable, and that whether the technology improves organizational sensing depends on the listening structure around the interface, not on the interface alone.

## 2. Theoretical Background

Four streams inform the study. We review each briefly and then state the gap.

**Employee voice and silence.** Voice is the discretionary communication of work-related ideas or concerns with intent to improve organizational functioning (Morrison, 2011); prohibitive voice, which names problems, carries higher interpersonal risk than promotive voice (Liang et al., 2012). Silence is the deliberate withholding of such input, in defensive and acquiescent forms (Morrison & Milliken, 2000; Van Dyne et al., 2003). Employees withhold because speaking up to those who evaluate them is risky: implicit voice theories suppress critique even absent sanction (Detert & Edmondson, 2011; Kish-Gephart et al., 2009), and futility, the expectation that input changes nothing, silences even when people feel safe (Morrison, 2014). Across this literature, silence is conceptualized as non-expression: the concern exists, but no utterance carries it. Even where voice and silence are treated as distinct constructs rather than poles of one continuum (Sherf et al., 2021), both remain located at the level of *whether* something is said, and measurement inherits the assumption that a favorable score marks the absence of a concern. Whether silence can also operate *inside* an expression has not been examined.

This is where our construct, *silence inside expression*, must be distinguished from adjacent ideas. It is not hedging, impression management, or the mum effect (reluctance to transmit bad news upward), nor upward message distortion alone, nor Hewlin's facade of conformity (Hewlin, 2003), in which employees suppress private values to feign agreement across a relationship over time. The novelty here is narrower and more observable: the co-occurrence, within a single episode, of an articulated concern and a favorable rating on the same theme across two channels, observed under real stakes. The employee neither stays silent nor feigns agreement wholesale; they voice the concern and simultaneously render a rating that does not carry it.

**Channel affordances.** Channel properties, and what employees believe them to be, shape voice independently of motivation (Ellmer & Reichel, 2021); perceived anonymity and low visibility raise a channel's safety for prohibitive voice (Mao & DeAndrea, 2019). Channels also differ in the *form* expression can take. A fixed rating cannot carry the qualification, attribution, and context that make a risky concern speakable, which makes it a low-cost place to keep a concern; conversation relaxes this, reducing measurement error (Schober & Conrad, 1997) and eliciting longer, more substantive answers from LLM-based interviewers (Wuttke et al., 2025). Channel comparisons should therefore expect differences not merely in how much is disclosed but in what form.

**Self-disclosure to conversational AI.** People disclose more honestly to agents believed to be non-human, reporting lower fear of evaluation (Ho et al., 2018; Lucas et al., 2014), with non-judgment and trust as central mediators (Papneja & Yadav, 2025). The effect has limits: people feel less heard once a response is labeled AI (Yin et al., 2024), and anthropomorphic agents can provoke unease (Van Zelderen et al., 2025). Field evidence is thin. Chatbots can function as voice channels (Dutta et al., 2023), but designs compare across persons or channels rather than within them, and most evidence comes from low-stakes settings.

**Organizational listening.** What a channel elicits is bounded by the listening structure: who attends to voice, how it is processed, and whether it produces a response (Yip & Fisher, 2022). Listening yields relational and performance benefits (Kluger & Itzchakov, 2022), but voice systems often fail at the response stage, reinforcing futility (Wilkinson et al., 2018). Getting employees to speak and earning their trust that speaking matters are thus different achievements.

**Research gap.** Read together, the streams assign complementary roles: voice and silence research explains *why* hierarchy-sensitive concerns stay compressed; channel and AI-disclosure research explains *how* a conversational channel might make them speakable; listening research explains *what* bounds candor even where a channel succeeds. None provides field evidence on how one concern, held by one employee in one week, is rendered by a structured survey and an adaptive AI conversation, or on the protective form disclosure takes. This study addresses that gap with a single-session, two-channel field design.

# 3. Method

## 3.1. Research design

This study pursues two goals: to characterize how employees disclose hierarchy-sensitive concerns to a conversational AI agent while registering favorable ratings on the same theme in a co-occurring survey, and to identify when that candor expands or contracts. Because these are "how" and "when" questions aimed at surfacing concepts and relationships rather than testing prior hypotheses, we adopt an inductive design and follow the Gioia methodology for rigor in inductive theory building (Gioia et al., 2013).

Gioia-style analysis assumes researcher-conducted interviews in which informants co-construct accounts with the researcher (Gioia et al., 2013); here, informants spoke to an AI agent deployed by their employer. Conducting interviews at scale through conversational agents is itself an emerging methodological approach in organizational research (Tamme et al., 2026). The transcripts are workplace disclosures to an organizational instrument, and the exposure management we theorize is precisely a response to being heard by the organization rather than by a neutral interviewer. We therefore read them as records of how employees manage disclosure to a corporate listening channel, and confine our claims to that object. As patterns emerged, we moved abductively between the data and the voice, channel, and AI-disclosure literatures, yielding a grounded model and propositions.

## 3.2. Setting and listening system

The study was conducted at a global management consulting firm whose hierarchies, high performance pressure, up-or-out advancement, and strong norms of upward deference may give employees incentives to compress critical upward concerns, making consulting a theoretically appropriate site for observing how candor is managed (Detert & Edmondson, 2011).

The introduced weekly listening process pairs two channels in one session. First, a short **structured pre-survey** asks employees to respond to fixed agreement items on a **four-point forced-choice scale** (1 = Disagree, 2 = Tend to disagree, 3 = Tend to agree, 4 = Agree; the scale has no neutral midpoint and no "strongly agree" option). Second, an **adaptive voice interview** conducted by a conversational AI agent covers five fixed themes: overall experience, workload sustainability, client value, leadership engagement, and one open topic. Three of these (workload sustainability, client value, and leadership engagement) have directly matching pre-survey items and form the basis of the channel-matched analysis; overall experience and the open topic have no matched survey item and are excluded from gap analysis.

The voice agent is built on the commercial conversational-AI platform Sona8 (YC, F26), which integrates automatic speech recognition (ASR), a large language model for dialogue management, and text-to-speech in a real-time spoken loop.

Rather than reading a fixed script, the agent probes adaptively: a theme-level system prompt instructs it to open each theme with a single open question, follow up on initial responses, ask for elaboration and specifics, and avoid leading or evaluative language, lowering the cost of articulating concerns a fixed rating cannot capture (Schober & Conrad, 1997; Wuttke et al., 2025). Sessions in our corpus averaged 6.05 minutes and 10.6 conversational turns, a bounded turn budget that ends a theme once the agent's follow-ups are satisfied or the respondent declines to elaborate. The platform produces a verbatim transcript via its built-in ASR, retaining natural speech features such as hesitations and false starts. Classifications were based on substantive wording and did not depend on disfluencies or incomplete utterances. Source audio was not retained for privacy reasons and therefore could not be independently verified.

## 3.3. Data, consent, and ethics

The first-session sample used for the channel-matched analysis comprises 44 unique interview sessions. Each pairs one employee's pre-survey responses with the transcript of their AI conversation, yielding 132 matched data points across the three matched themes. To preserve independence and avoid conditioning effects from repeated exposure to the agent, we restrict the analysis to each participant's first completed session. Sessions that were thin or technically compromised (early dropouts, severe truncation) were excluded under documented rules; two first-session transcripts fell below the length-and-substance threshold, leaving 44, and because both contained no gap the exclusion does not affect the gap count. By contrast, the Gioia analysis draws on all 65 eligible interview sessions to identify recurring forms of disclosure and develop conceptual categories, yielding 158 mechanism-bearing quotes. Retained transcripts ranged from 58 to 1028 words (M = 414, SD = 263.59). Analyses operate at two levels: gap prevalence is measured across the 132 matched theme-sessions from first eligible sessions, while disclosure mechanisms are analyzed across 158 protective quotes from all 65 eligible interview sessions.

The listening program was deployed by the employer, and all employees took part voluntarily and were informed that anonymized transcripts could be used for research. Personal names, client names, and other identifying references were removed or masked, and only session-level pseudonymous identifiers were retained. Voluntariness inside an employer-deployed tool is not equivalent to neutral research consent, since employees engaged a channel their organization supports; we therefore report aggregate patterns and short illustrative quotations rather than session narratives.

### 3.4. Analytic procedure

Our analysis proceeds in two linked steps: identifying where survey and conversation diverge, then theorizing how that divergence is produced.

**Operationalizing the gap.** We define a *disclosure-gap episode* as a favorable Likert rating on a theme co-occurring with a substantive concern on that same theme in the conversation. We treated a rating as **favorable (nonnegative) when it fell on the agree side of the four-point scale**, the top two boxes "Tend to agree" or "Agree", and a concern as substantive when the transcript contained an explicit problem, complaint, or unmet need on that theme rather than a neutral or descriptive remark. Each theme-level episode was classified for disclosure level (absent, hedged, or explicit) and assigned a gap level: *none* where rating and disclosure aligned, *moderate* where a favorable rating co-occurred with a hedged or qualified concern, and *strong* where a favorable rating co-occurred with an explicit concern. A top-of-scale workload rating paired with an account of repeated late nights, for instance, was coded as a strong gap. The theme-matching rule was conservative: a survey item and a conversational segment were treated as the same theme only for the three items with one-to-one correspondence between a pre-survey item and a fixed interview theme.

**Reliability.** Two coders independently assigned gap levels to a double-coded set of 132 theme-session ratings (both coders' full pass across the matched themes). On the binary gap/no-gap decision that carries the findings, agreement was 81.0%, Cohen's $\kappa$ was 0.47 and the prevalence-adjusted bias-adjusted $\kappa$ PABAK = 0.62 (Byrt et al., 1993). On the three-level scale, raw agreement was 78.0% and $\kappa$ was 0.34 unweighted, rising to 0.44 with quadratic weights once the ordering of none, moderate, and strong is credited. The unweighted $\kappa$ is depressed by the heavily skewed marginal distribution, the prevalence effect by which high agreement coincides with a conservative $\kappa$ (Feinstein & Cicchetti, 1990); we therefore report raw agreement and weighted $\kappa$ alongside it. Disagreements were resolved by discussion before the analysis reported below.

**Theorizing the gap.** Following the Gioia methodology (Gioia et al., 2013), we developed first-order codes in participants' language, abstracted them into second-order themes naming the mechanisms through which employees disclose yet protect themselves, and grouped these into aggregate dimensions. A large language model, given only anonymized transcripts, surfaced first-pass candidate quotes and codes with confidence scores, but all review, final coding decisions, and interpretive abstraction were performed by the authors. Consistent with evidence that such models reproduce concrete, descriptive themes more reliably than subtle interpretive ones (Morgan, 2023), we used the model strictly as a proposal mechanism. In a random audit of 15 sessions (23% of the corpus), 49 of 50 model-proposed quote/code pairs survived human review (98%), while the model surfaced 45 of 46 passages independently identified by the author (98% coverage). No second-order theme or aggregate dimension originated from an LLM suggestion; these were developed by the authors from the human-reviewed first-order material. This provides an additional safeguard against circularity, since an LLM both elicited the disclosures and helped surface candidate codes. We did not compute inter-rater statistics for the interpretive second-order layers; rigor there comes from the traceability of the data structure (Gioia et al., 2013). The descriptive mechanism counts reported below characterize the qualitative corpus and are not statistical inferences (Maxwell, 2010); the study's claims rest on the grounded model and propositions.

## 4. Findings

### 4.1. Disclosure-gap episodes

If silence operated only as non-expression, a favorable survey rating would signal the absence of a concern. It does not. Across the 44 sessions, each matched on three themes, **41% of sessions (18 of 44)** contained at least one favorable rating that co-occurred with a substantive concern voiced on the same theme in the same session. At the level of individual ratings, **18% of theme-sessions (24 of 132)** contained such a disclosure gap. Silence inside expression is therefore, in this corpus, not a marginal case but a recurring feature of the survey channel.

The gaps surface across all three matched themes, each carrying upward risk in a professional-services setting: workload critiques implicate the leaders who set staffing and deadlines, client-value doubts question engagements, and leadership feedback is openly evaluative of superiors. Gaps appeared in 20% of leadership ratings (9 of 44), 18% of workload ratings (8 of 44), and 16% of client-value ratings (7 of 44), present throughout rather than confined to one topic. Leadership engagement shows the gap most often, consistent with its being the most directly hierarchy-implicating of the three.

**Robustness to the favorability threshold.** Because the headline depends on what counts as favorable, we re-ran prevalence under a stricter cut. On this four-point scale there is no "5," so the strictest threshold is the top box alone ("Agree"); Table 1 reports both. The phenomenon holds in kind: even when only an unqualified "Agree" counts as favorable, one session in five still hides a voiced concern. It is sensitive in degree, since counting the agree-side "Tend to agree" as favorable roughly doubles prevalence. We adopt the top-two-box threshold and report the top-box figure as a conservative lower bound.

**Table 1. Gap prevalence by favorability threshold**

| Threshold | Theme-sessions w/ gap | Sessions w/ gap |
|---|---|---|
| Top-two-box (Tend to + Agree) | 24/132 (18%) | 18/44 (41%) |
| Top box only (Agree) | 11/132 (8%) | 9/44 (20%) |

**A rival explanation.** An agent that probes for specifics will elicit problem-talk, so the divergence could reflect question form rather than concealed silence (Schober & Conrad, 1997). But the agent opened every theme with the same neutral question and probed identically for all participants (§3.2), so probing raises what is sayable for everyone equally; it cannot explain why, for some, a raised concern is compressed beneath a favorable rating while for others it is not. The difference lies in the *kind* of protective work employees do, which we quantify in §4.3.

### 4.2. The structure of disclosure

When employees raise concerns in the AI conversation, they rarely disclose them outright; disclosure and self-protection co-occur in the same utterance. The data structure organizes this protective work into ten second-order themes and four aggregate dimensions (Figure 1), drawn from 158 mechanism-bearing quotes.

The four dimensions describe distinct ways employees disclose yet protect themselves. **Softening the disclosed concern (45% of quotes)** is the most frequent protective move: employees voice a concern but cushion it through hedging, by normalizing extreme conditions, or by recasting a complaint as a wish. **Deflecting accountability (37%)** redirects responsibility for a problem away from named superiors toward the client or circumstance, or the speaker disclaims the standing to judge. **Bounding disclosure depth (11%)** limits how far a concern is taken, pulling back when probed or naming an issue while declining to escalate it. **Conditional legitimacy of AI listening (7%)** ties disclosure to the channel itself, surfacing doubt that speaking up will matter. Each dimension is grounded in participant language (Table 2).

Softening is not the most frequent protective move overall. As §4.3 shows, its descriptive distribution differs between gap and no-gap sessions, whereas deflection occurs at similar rates across the two groups. Softening is the move that varies with the gap, which is why we treat it as particularly informative for understanding how a concern can sit beneath a favorable rating.

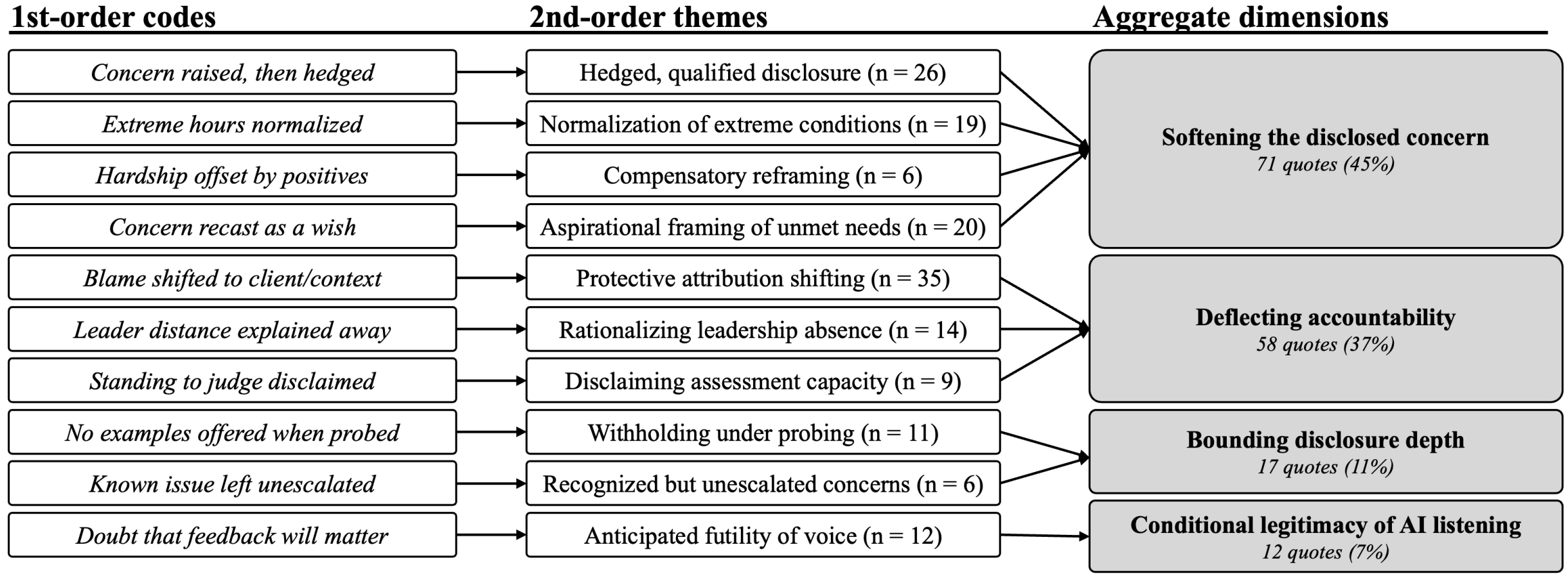

**Figure 1. Disclosure-mechanism data structure (All eligible sessions; N = 158 protective quotes).**

**Table 2. Aggregate dimensions with illustrative participant quotes**

| Aggregate dimension (share) | Illustrative participant quotes |
|---|---|
| Softening the disclosed concern (45%) | "not 100% sure if it is sustainable… working really long hours"; long hours "felt sustainable because of the good team" |
| Deflecting accountability (37%) | "due to the time difference to the US… the stakeholders had…"; "several projects running… harder to engage fully" |
| Bounding disclosure depth (11%) | "Nothing comes to mind for concrete examples"; a known crunch "was not good" yet deliberately not pushed |
| Conditional legitimacy of AI listening (7%) | "I hope… they will take into account the feedback we gave" |

## 4.3. Two signatures of silence

Descriptively, the distribution of protective mechanisms differs between gap and no-gap material in this corpus (Table 3). Reading across the two groups, two patterns inform the grounded model.

**Softening is particularly prominent in gap material.** They voice the concern but cushion it as they go, so it never reaches the weight a rating would register. A participant naming very late nights frames them as workable, or as already improving; another offsets long hours against a good team atmosphere. Softening accounts for 35 of 59 protective quotes in gap sessions (59%), against 36 of 99 (36%) where no gap forms, and constitutes 64% of protective quotes in strong-gap sessions. This pattern is consistent with the grounded interpretation that qualification can allow a voiced concern to coexist with a favorable score.

**Where no gap forms, the protective work looks different**, and this is the more telling pattern because it is not built into how we defined the gap. Deflection runs at essentially the same level in both kinds of session (20 of 59, 35%, in gap sessions; 38 of 99, 38%, where none forms). Its similar prevalence across the two groups suggests that, in this corpus, deflection does not distinguish the two patterns. Bounding disclosure depth and conditional legitimacy of AI listening are more prevalent in no-gap material: together, they account for 25 of 99 protective quotes in no-gap sessions (25%), compared with 4 of 59 in gap sessions (7%).

**Table 3. Protective mechanisms by session gap**

| Dimension | Gap quotes (n = 59) | No-gap quotes (n = 99) |
|---|---|---|
| Softening the disclosed concern | 35 (59%) | 36 (36%) |
| Deflecting accountability | 20 (35%) | 38 (38%) |
| Bounding disclosure depth | 3 (5%) | 14 (14%) |
| Conditional legitimacy of AI listening | 1 (2%) | 11 (11%) |

These are two forms of silence. The gap shows silence operating *inside* expression: the concern is spoken but softened below what a rating can detect. The no-gap sessions point to silence in its familiar sense, the concern held back or foreclosed by the belief that voicing is pointless. A favorable rating with no disclosed concern, then, does not always mean agreement; sometimes the concern never surfaced at all. The same futility that keeps it down also limits what the channel can draw out, which is where perceived legitimacy, developed next, begins.

## 5. A Grounded Model of Bounded Disclosure

The data structure (Figure 1) specifies the concepts through which employees disclose yet protect themselves; the process model in **Figure 2** shows how they relate in motion (Gioia et al., 2013) to produce the central puzzle: the same concern registering as a favorable rating in one channel and a managed disclosure in the other. We develop the model in four moves, each with a proposition. Because the design is cross-sectional and observational, we phrase these as associations to be tested, not established effects.

### 5.1. From hierarchical risk to compressed ratings

The model begins with the contextual conditions of the setting. Steep hierarchy, up-or-out advancement, and evaluative pressure make prohibitive concerns about workload, leadership, and engagements costly to voice upward (Detert & Edmondson, 2011). When such a concern arises, the structured survey offers no place to put it safely: a fixed rating cannot carry the qualification, attribution, and context through which our participants made concerns speakable, only a position on a scale the employee knows will be read and aggregated. The favorable rating that results is, on this reading, less a measurement error than an act of exposure management, and the survey functions as a safe compression device.

**Proposition 1.** *Under conditions of hierarchical risk, prohibitive concerns are more likely to surface in an adaptive AI conversation than in a co-occurring structured-survey rating on the same theme, such that favorable ratings under-represent the prevalence of concerns.*

### 5.2. Probing lowers the cost of articulation, but disclosure arrives managed

The conversational channel changes what the rating could not: by probing adaptively for specifics, the agent lowers the cost of articulating a concern that requires explanation to be voiced at all (Schober & Conrad, 1997; Wuttke et al., 2025). Yet across the 158 mechanism-bearing quotes, disclosure and self-protection co-occur in the same utterance. The channel changes what is sayable; it does not remove exposure management. From here the model branches into two pathways, the two signatures of silence in §4.3.

### 5.3. Pathway A: Softening produces the disclosure gap

Where a gap forms, disclosure travels through softening. The concern is expressed gently, hedged, balanced with a positive counter, or rephrased as a hope, such that it does not carry the weight a rating would imply (59% of protective quotes in gap sessions, rising to 64% in strong-gap sessions; §4.3). This is silence operating inside expression: the concern reaches the channel, but its force is managed below the threshold of the co-occurring rating. Deflection, by contrast, runs at similar rates whether or not a gap forms, marking it a baseline attribution habit rather than a pattern specific to compression.

**Proposition 2.** *Where disclosure gaps form, disclosure is carried predominantly by softening mechanisms: the more a concern is hedged, normalized, or compensatorily reframed in conversation, the more likely it is to co-occur with a favorable rating on the same theme.*

### 5.4. Pathway B: Foreclosure produces silence without a gap

Where no gap forms despite a favorable rating, the protective work looks different. Employees on this pathway stop short when probed, decline to escalate concerns they name as known, or foreclose disclosure with the anticipation that voicing will change nothing (bounding and futility together 25% of no-gap protective quotes, against 7% in gap sessions). The implication inverts the usual reading of survey data: a favorable rating without a disclosed concern does not certify the absence of a concern. Sometimes it marks silence in its classic form (Morrison, 2014).

**Proposition 3.** *Favorable ratings unaccompanied by disclosed concerns do not reliably indicate the absence of concerns; in sessions without disclosure gaps, silence is more likely to operate through withholding under probing and anticipated futility of voice than through softened expression.*

### 5.5. Legitimacy as the conditioning boundary

Both pathways appear conditioned by the perceived legitimacy of the listening structure itself.

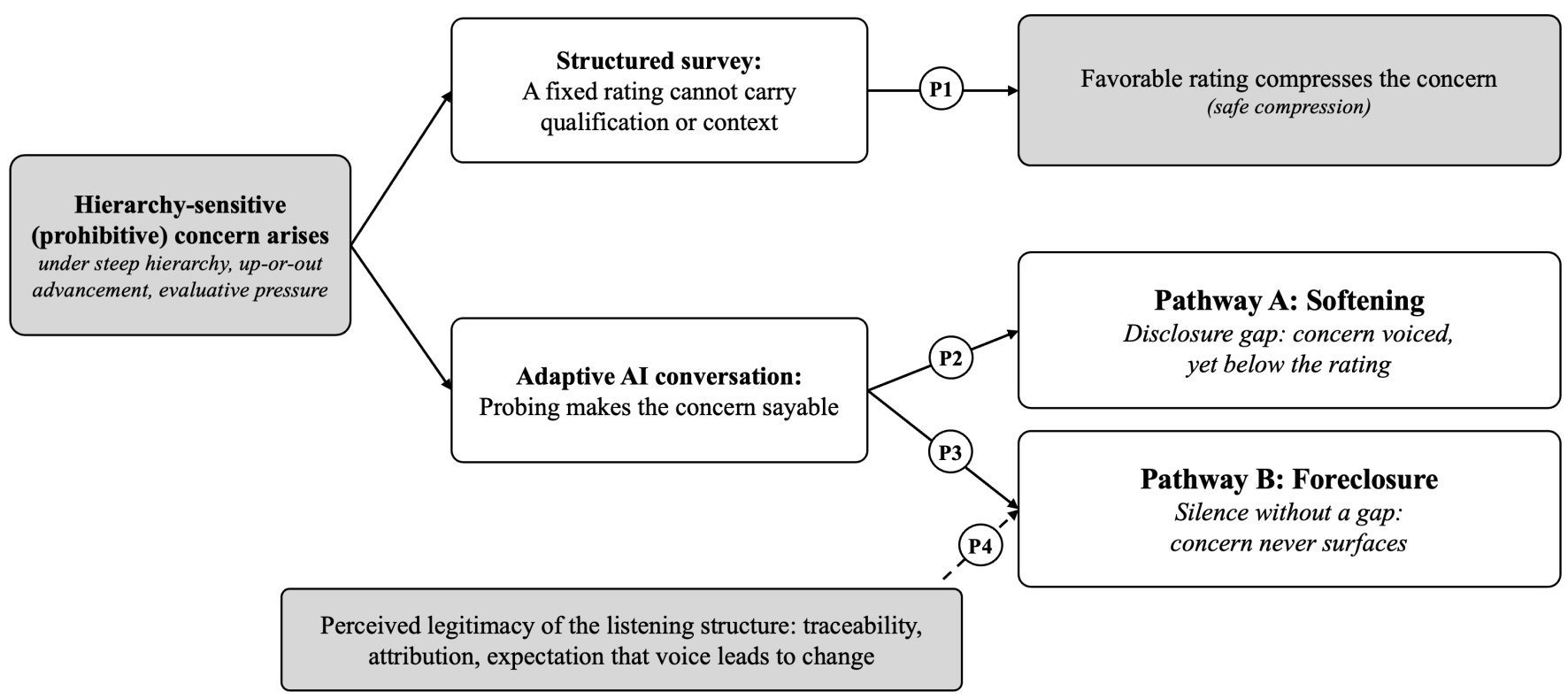

**Figure 2. Grounded process model of bounded disclosure**

Participants tied their candor to the channel: requesting confidentiality for what was just shared, treating openness to the AI as separate from the work routine, and, most consequentially, doubting that anything said would lead to change. Legitimacy concerns were rarer in gap sessions (1 of 59 protective quotes, 2%) than where no gap formed (11 of 99, 11%), and the broader foreclosure signature (bounding plus legitimacy) is far more concentrated where no gap forms (25% versus 7%). Where the channel is felt to be legitimate, modulation appears to relax toward softened disclosure; where its consequences are doubted, disclosure forecloses. Eliciting voice and earning trust in the listening structure are thus distinct achievements (Yip & Fisher, 2022): adaptive probing accomplishes the first, but only felt legitimacy seems to govern the second.

We flag Proposition 4 as the most tentative of the four. It rests on the thinnest evidence in our corpus, since the legitimacy dimension comprises just 12 quotes (7%), so we advance it as a conjecture for future test rather than a finding.

**Proposition 4 (conjecture).** *The perceived legitimacy of the listening structure bounds how far protective modulation relaxes: lower perceived legitimacy is associated with a shift from softened disclosure toward withholding and anticipated futility, capping the candor any conversational channel can sustain.*

## 6. Discussion

We asked when and how employees disclose hierarchy-sensitive concerns to a conversational AI agent while registering favorable ratings on the same theme in a co-occurring survey. Our grounded model holds that the two channels do not differ merely in how much they elicit, but in the form self-protection takes within each. In the survey, protection operates through compression: the concern is folded into a favorable rating. In the conversation, protection operates through modulation: the concern is voiced but softened, deflected, or bounded, and where the legitimacy of the channel is doubted, foreclosed. Candor, in short, is bounded rather than binary, and the boundary moves with the channel.

### 6.1. Theoretical contributions

First, we extend voice and silence theory by identifying a second locus of silence. The literature treats silence as non-expression, the withholding of concerns judged too risky to voice (Morrison, 2023; Milliken et al., 2003). Our no-gap sessions confirm this classic form (Proposition 3). The gap episodes, however, show that silence can also persist inside expression: in 41% of sessions a concern was spoken in conversation yet softened below the threshold the co-occurring rating registered (Propositions 1 and 2). An employee can thus be voicing and silent on the same concern at the same moment, depending on which instrument is listening. This divergence is not reducible to the two channels affording different output forms: probing was held constant across participants, yet the same favorable rating concealed a voiced concern for some and not others, pointing to employee exposure management rather than an instrument artifact. This is what distinguishes *silence inside expression* from facades of conformity, impression management, or the mum effect: a single concern rendered divergently across two co-occurring channels under real stakes. The measurement corollary is that the survey is better read as a reliable compression device than as a flawed instrument, so consistently favorable pulse scores mark a ceiling on expressed discontent rather than evidence of contentment.

Second, we contribute field evidence to research on self-disclosure to conversational agents, developed so far largely in laboratory and consumer settings (Lucas et al., 2014; Papneja & Yadav, 2025). Observing the same employee, theme, and week across two channels under real hierarchical stakes shows that the disclosure-enhancing pattern of conversational AI survives in the field, but in qualified form: instead of the uninhibited candor predicted by anonymity-based accounts, we observe *managed* disclosure, every expressed concern accompanied by protective work. The question for this literature is not whether people disclose more to AI, but what form their self-protection takes when they do.

Third, we offer a socio-technical reframing of AI-mediated listening as a contribution to research on technology and the future of work. Eliciting voice and earning trust in the listening structure are distinct achievements (Kluger & Itzchakov, 2022; Yip & Fisher, 2022). Adaptive probing accomplished the first, drawing out concerns that ratings concealed; the second appears governed by felt legitimacy, namely traceability, attribution, and the expectation that speaking leads to change (Proposition 4, advanced tentatively). Listening capability is thus a property not of the interface but of the loop: the channel determines what becomes sayable, while the organization's visible response determines what remains worth saying. For systems that sense a workforce at scale, these are separable design targets.

## 6.2. Implications for practice

For organizations deploying AI-mediated listening, the design problem is twofold. To surface what surveys compress, the channel must probe: adaptive follow-up lowered the cost of articulation in our setting, so agents should pursue specifics rather than collect open comments passively. Because disclosure arrives softened, deflected, and bounded, the analysis layer should read modulated signals rather than score sentiment at face value: a hedged account of late nights ("a bit longer from time to time") should raise, not lower, a workload flag, and managers should treat favorable pulse scores as an upper bound on contentment, not evidence of its absence (Proposition 1). Beyond the interface, organizations should manage the channel's *legitimacy* by communicating who sees what, at what level of attribution, and by visibly closing the feedback loop. Our data suggest reticence stemmed more from anticipated futility than from distrust of the AI itself (Proposition 4): a channel without visible follow-through risks the worst of both worlds, extracting more sensitive disclosures than a survey while reinforcing the belief that speaking up changes nothing.

## 6.3. Limitations and future research

Four limitations bound our claims. First, the study draws on a single professional-services firm whose steep hierarchy intensifies the phenomenon; the pathways may apply to other high-pressure settings, but mechanism weights will vary by context. Second, participation was voluntary and the channel was used repeatedly, so self-selection may overrepresent employees already willing to engage it, and the foreclosure pathway may be more prevalent than our data show. Third, the transcripts are not a gold standard against which the survey errs: the conversation changes the disclosure task itself, so the gap measures a *difference between channels*, not a distance from truth. Our central claim is therefore about channel divergence, not concealment of a known ground truth. Fourth, because source audio was not retained for privacy reasons, transcription accuracy could not be independently verified, and occasional interruptions might have truncated some disclosures. Our prevalence figures are descriptive counts meant to characterize the corpus, not to generalize statistically; the legitimacy dimension especially rests on few quotes.

These limitations chart the agenda. The propositions are testable: experimental manipulation of legitimacy cues, such as attribution transparency and visible loop-closing, would isolate their effect on the softening-to-foreclosure balance (Proposition 4), and cross-organizational comparison would test whether hierarchy steepness shifts it (Propositions 1 to 3). Longitudinal designs could test whether repeated exposure relaxes modulation as legitimacy is earned or deepens futility when follow-through fails, and linking disclosure gaps to outcomes such as turnover would show whether the concerns surveys compress are the ones organizations most need to hear.

# 7. Conclusion

Organizations increasingly listen to employees through AI, yet what such channels change about what employees actually say has remained open. Comparing the same employees, themes, and weeks across a structured survey and an adaptive AI conversation, we find that favorable survey ratings under-represent the concerns employees will articulate once a channel lets them; that this divergence is patterned, with softening prominent where gaps appear and foreclosure more prominent where they do not; and that it is bounded by the felt legitimacy of the listening structure. Where employees doubt that speaking leads to change, disclosure forecloses before any channel can reach it. A favorable score, in the end, tells an organization only that nothing was said. What can be said depends on the channel; what is worth saying depends on the organization.